\documentclass{llncs}

\usepackage[T1]{fontenc}

\usepackage{algorithm}
\usepackage{algpseudocode}

\usepackage{amsfonts}
\usepackage{amsmath}
\usepackage{tikz}
\usetikzlibrary{math} 
\usepackage{hhline}

\newcommand{\oof}{$1/f$ }

\begin{document}
\title{PEPS: Positional Encoding Projected Sampling - Extended }
\author{Guillaume Perez \and Janarbek Matai \and Takahiro Harada}
\institute{Advanded Micro Devices}



\maketitle

\begin{abstract}
Implicit neural representations (INRs) are increasingly being used as tools to map coordinates to signals, 
encompassing applications from neural fields to texture compression, shape representations, and beyond. 
Most INR methods are based on using high-dimensional projections of the initial coordinates through encoders such as grid or positional encoding.
Nevertheless, positional encoding is often insufficient 
and grids, as we show in this paper, require high resolution for being able to learn.
In this paper, we demonstrate that positional encoding can be used not only as a high-dimensional embedding 
but also decomposed as a series of meaningful points. 
We propose the Positional Encoding Projected Sampling, 
where we treat the projection of the original coordinate at each frequency as a point of interest. 
We describe the motion of each point with respect to the frequencies and show that it follows a unique pattern.
Finally, we use the unique motion of each point as a basis decomposition for doing learned positional encoding using grids.
We prove, using three competitive applications— image representation, texture compression, and signed distance function—
that the proposed approach outperforms the current state of the art methods, 
and often requires 25\% less parameters for equivalent reconstruction error or rendering.
\end{abstract}

\begin{figure*}
    \centering
    \begin{tikzpicture}[scale=0.8]
    
	\draw[draw=black, thin, solid]  (0.00,4.00) grid (4.00,0.00);
	\draw[draw=black, thin, solid] (1.00,3.00) circle (0.1);
	\draw[draw=black, thin, solid] (2.00,3.00) circle (0.1);
	\draw[draw=black, thin, solid] (3.00,3.00) circle (0.1);
	\draw[draw=black, thin, solid] (4.00,3.00) circle (0.1);
	\draw[draw=black, thin, solid] (4.00,4.00) circle (0.1);
	\draw[draw=black, thin, solid] (3.00,4.00) circle (0.1);
	\draw[draw=black, thin, solid] (2.00,4.00) circle (0.1);
	\draw[draw=black, thin, solid] (1.00,4.00) circle (0.1);
	\draw[draw=black, thin, solid] (0.00,4.00) circle (0.1);
	\draw[draw=black, thin, solid] (0.00,3.00) circle (0.1);
	\draw[draw=black, thin, solid] (0.00,2.00) circle (0.1);
	\draw[draw=black, thin, solid] (1.00,2.00) circle (0.1);
	\draw[draw=black, thin, solid] (2.00,2.00) circle (0.1);
	\draw[draw=black, thin, solid] (3.00,2.00) circle (0.1);
	\draw[draw=black, thin, solid] (4.00,2.00) circle (0.1);
	\draw[draw=black, thin, solid] (4.00,1.00) circle (0.1);
	\draw[draw=black, thin, solid] (3.00,1.00) circle (0.1);
	\draw[draw=black, thin, solid] (2.00,1.00) circle (0.1);
	\draw[draw=black, thin, solid] (1.00,1.00) circle (0.1);
	\draw[draw=black, thin, solid] (0.00,1.00) circle (0.1);
	\draw[draw=black, thin, solid] (0.00,0.00) circle (0.1);
	\draw[draw=black, thin, solid] (1.00,0.00) circle (0.1);
	\draw[draw=black, thin, solid] (2.00,0.00) circle (0.1);
	\draw[draw=black, thin, solid] (3.00,0.00) circle (0.1);
	\draw[draw=black, thin, solid] (4.00,0.00) circle (0.1);
    
    \tikzmath{\x = 8; \y = 4; } ;
    \filldraw[cyan]  ({\x/10},{\y/10})  circle (1pt) node[anchor=south east] {$x$};
    \foreach \z in {1,2,3}
        {
          \filldraw[purple] ({2*sin(\x*2^\z*pi)+2},{2*sin(\y*2^\z*pi)+2}) circle (1pt) node[anchor=west]{$S_{\phi_{\z}}$};
          \filldraw[orange] ({2*cos(\x*2^\z*pi)+2},{2*cos(\y*2^\z*pi)+2}) circle (1pt) node[anchor=west]{$C_{\phi_{\z}}$};
        }

    \draw[draw=cyan, -latex, thin, dashed] ({\x/10-5},{\y/10}) .. controls ({\x/10-3},{\y/10+0.2}) and ({\x/10-1},{\y/10+0.2}) .. ({\x/10},{\y/10});
     \tikzmath{\z = 3; } ;
    \draw[draw=orange, -latex, thin, dashed] ({2*cos(\x*2^\z*pi)+2-5},{2*cos(\y*2^\z*pi)+2}) .. controls ({2*cos(\x*2^\z*pi)+2-3},{2*cos(\y*2^\z*pi)+2.2}) and ({2*cos(\x*2^\z*pi)+2-1},{2*cos(\y*2^\z*pi)+2.2}) .. ({2*cos(\x*2^\z*pi)+2},{2*cos(\y*2^\z*pi)+2});
    
    \draw[draw=black, thin, solid]  (-5.00,4.00) rectangle (-1.00,0.00);
    \filldraw[cyan]  ({\x/10-5},{\y/10})  circle (1pt) node[anchor=south east] {$x$};
    \foreach \z in {1,2,3}
        {
          \filldraw[purple] ({2*sin(\x*2^\z*pi)+2-5},{2*sin(\y*2^\z*pi)+2}) circle (1pt) node[anchor=west]{$S_{\phi_{\z}}$};
          \filldraw[orange] ({2*cos(\x*2^\z*pi)+2-5},{2*cos(\y*2^\z*pi)+2}) circle (1pt) node[anchor=west]{$C_{\phi_{\z}}$};
        }
    \foreach \z in {0,1,...,1500}
        {
            \draw[draw=purple] ({2*sin(\x*\z)+2-5},{2*sin(\y*\z)+2}) -- ({2*sin(\x*(0.5+\z))+2-5},{2*sin(\y*(0.5+\z))+2});
            \draw[draw=orange]  ({2*cos(\x*\z)+2-5},{2*cos(\y*\z)+2}) -- ({2*cos(\x*(0.5+\z))+2-5},{2*cos(\y*(0.5+\z))+2});
        }

    \tikzmath{\s = 0.25; } ;
    \tikzmath{\c = 5.50; \v =2.60; } ;
    \draw[draw=black, fill=cyan, thin, solid] ({\c},{\v}) rectangle ({\c+\s},{\v-\s});
    \foreach \z in {1,2,3}
        {
	\draw[draw=black, fill=purple, thin, solid] ({\c},{\v-\s*\z}) rectangle ({\c+\s},{\v-\s*(\z+1)}) ;
    }
    \foreach \z in {4,5,6}
        {
	\draw[draw=black, fill=orange, thin, solid] ({\c},{\v-\s*\z}) rectangle ({\c+\s},{\v-\s*(\z+1)}) ;
    }

    \tikzmath{\z = 3; };
    \draw[draw=cyan, -latex, thin, dashed] ({\x/10},{\y/10}) .. controls ({\x/10+1},{\y/10-0.5}) and ({\x/10+4},{\v-0.2}) .. ({\c-0.1},{\v-0.1});
    \draw[draw=orange, -latex, thin, dashed] ({2*cos(\x*2^\z*pi)+2},{2*cos(\y*2^\z*pi)+2}) .. controls ({2*cos(\x*2^\z*pi)+2+3},{2*cos(\y*2^\z*pi)+2-0.5})  .. ({\c-0.1},{\v-\s*(\z+3)-0.1});

    \tikzmath{\dec = \c+1.5; \sepL = 1;};
    \foreach \z in {0,1,2}
        {
            \foreach \d in{0,1,2,3,4}
            {
    	   \draw[draw=black] (\dec,\v +0.5-\z) -- (\dec+ \sepL,\v +1.5 - \d);
            }
        }
    \foreach \z in {0,1,2}
        {
            \foreach \d in{0,1,2,3,4}
            {
    	   \draw[draw=black] (\dec+ 3*\sepL,\v +0.5-\z) -- (\dec+ 2*\sepL,\v +1.5 - \d);
            }
        }

    \draw[draw=black, fill=cyan, thin, solid] (\dec,\v +0.5) ellipse (0.15 and 0.15);
    \draw[draw=black, fill=purple, thin, solid] (\dec,\v -0.5) ellipse (0.15 and 0.15);
    \draw[draw=black, fill=orange, thin, solid] (\dec,\v -1.5) ellipse (0.15 and 0.15);
    
    \foreach \z in {0,1,2,3,4}
        {
            \draw[draw=black, fill=black, thin, solid] (\dec+ \sepL,\v +1.5 - \z) ellipse (0.15 and 0.15);
        }
    \foreach \z in {0,1,2,3,4}
        {
            \draw[draw=black, fill=black, thin, solid] (\dec+ 2*\sepL,\v +1.5 - \z) ellipse (0.15 and 0.15);
        }
        
    \draw[draw=black, fill=red, thin, solid] (\dec+ 3*\sepL,\v +0.5) ellipse (0.15 and 0.15);
    \draw[draw=black, fill=green, thin, solid] (\dec+ 3*\sepL,\v -0.5) ellipse (0.15 and 0.15);
    \draw[draw=black, fill=blue, thin, solid] (\dec+ 3*\sepL,\v -1.5) ellipse (0.15 and 0.15);

    \draw[draw=black, fill=black, thin, solid] (\dec+\sepL*1.3,\v -0.5) ellipse (0.05 and 0.05);
    \draw[draw=black, fill=black, thin, solid] (\dec+\sepL*1.5,\v -0.5) ellipse (0.05 and 0.05);
    \draw[draw=black, fill=black, thin, solid] (\dec+\sepL*1.7,\v -0.5) ellipse (0.05 and 0.05);

    \tikzmath{\z = 3; };
    \draw[draw=cyan, -latex, thin, dashed] ({\c+\s+0.1},{\v-0.1}) -- ({\dec -0.1},{\v +0.5});
    \draw[draw=orange, -latex, thin, dashed] ({\c+\s+0.1},{\v-\s*(\z+4)+\s-0.1}) -- ({\dec-0.1},{\v -1.5});

    \tikzmath{\step = 0.1; };
    
    \node[inner sep=0pt] (diff) at (\dec+ 5.5*\sepL,\v -.5)
    {\includegraphics[width=.1\textwidth]{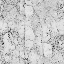}};

    \node[inner sep=0pt] (diff) at (\dec+ 5.5*\sepL - \step * 1,\v -.5+ \step * 1)
    {\includegraphics[width=.1\textwidth]{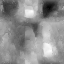}};
    
    \node[inner sep=0pt] (diff) at (\dec+ 5.5*\sepL - \step * 2,\v -.5+ \step * 2)
    {\includegraphics[width=.1\textwidth]{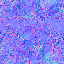}};

    \node[inner sep=0pt] (diff) at (\dec+ 5.5*\sepL - \step * 3,\v -.5+ \step * 3)
    {\includegraphics[width=.1\textwidth]{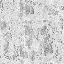}};
    
    \node[inner sep=0pt] (diff) at (\dec+ 5.5*\sepL - \step * 4,\v -.5+ \step * 4)
    {\includegraphics[width=.1\textwidth]{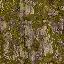}};

    \tikzmath{\z = 3; };
    \draw[draw=red, thin, dashed] (\dec+ 3*\sepL,\v +0.5) -- (\dec+ 5.5*\sepL - \step * 4-0.5,\v -.5+ \step * 4-0.7);
    \draw[draw=green, thin, dashed] (\dec+ 3*\sepL,\v -0.5) -- (\dec+ 5.5*\sepL - \step * 4-0.5,\v -.5+ \step * 4-0.7);
    \draw[draw=blue, thin, dashed] (\dec+ 3*\sepL,\v -1.5) -- (\dec+ 5.5*\sepL - \step * 4-0.5,\v -.5+ \step * 4-0.7);

    \node at ({-3},{-0.5})  {(a)};
    \node at ({2},{-0.5})  {(b)};
    \node at ({5.6},{-0.5})  {(c)};
    \node at ({8.5},{-0.5})  {(d)};
    \node at ({12.3},{-0.5})  {(e)};
    
\end{tikzpicture}
    \caption{Grid-PEPS method applied to neural texture compression of a texture set of size $k$.
    (a) Given a coordinate point $x\in \mathbb{R}^2$, the method extracts $S_{\phi}={\sin{x\phi}}$ and $C_{\phi}={\cos{x\phi}}$ for all the 
    targeted frequencies along the frequency curves.
    (b) Then for each point in $P=(x,S_{\phi_1},\dots,C_{\phi_k})$ it
    samples the grid to get latent values.
    (c) The result is the concatenation of the latent vector of each point. 
    (d) The resulting vector is fed to the neural network.
    (e) The output of the neural network $T(x) \in \mathbb{R}^{3k}$ is the values of the textures of the set at coordinate $x$.
    The pseudo-code is given by Algorithm~\ref{alg:pinkpecs} without the pink part.}
    \label{fig:gridpecs}
\end{figure*}
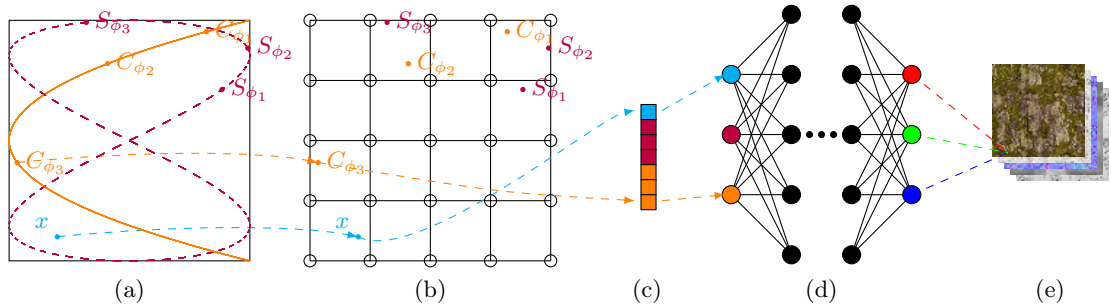
\section{Introduction}
Implicit neural representations (INRs) have emerged as a powerful tool for modeling multi-dimensional signals. 
INRs are models that learn coordinate-to-signal functions using neural networks. 
They are used in various domains, ranging from computer graphics—where they capture two-dimensional textures for texture compression \cite{nvidNTC23,farhadzadeh2024neural} or three-dimensional signed distance functions for rendering and NeRF \cite{mildenhall2021nerf,fujieda2024neural} to inverse problems and signal denoising \cite{saragadam2023wire}.
Most INRs are based on multi-layer perceptrons (MLPs), in which the activation functions, 
the input embedding, and other components have been dedicated to handling the low-dimensional coordinate input signal. 
The reason for these dedicated components is the spectral bias 
 of MLPs, which often struggle to represent high-frequency signals from low-dimensional inputs \cite{rahaman2019spectral}.

The best-known INR method to tackle this bias is to apply absolute positional encoding 
to the low-dimensional coordinates to project them into a higher-dimensional space \cite{tancik2020fourier,mildenhall2021nerf}.
Analogous to the positional encoding used in large language models (LLMs) \cite{vaswani2017attention}, 
INR positional encoding applies various multiplicative coefficients to the coordinates before applying a sinusoidal function.
The greater the number of coefficients, the greater the resulting number of dimensions.
This positional encoding, combined with a simple MLP, 
has been shown to be an efficient way to project low-dimensional coordinates into a high-dimensional space 
and to learn many coordinate-to-signal functions.
Recently, various methods have been developed to increase the representational capabilities of INRs, 
for example, by replacing ReLU activations in MLPs with periodic functions \cite{sitzmann2020implicit,ramasinghe2022beyond,saragadam2023wire}. 
Most of these works come at the cost of sensitivity to initial parameters, longer training times, and, most importantly, longer inference times, preventing their use in real-time applications.

To overcome this, several applications of INRs, such as texture compression in graphics, use learned positional encoding. 
In learned positional encoding, a vector of parameters is learned for each coordinate during training.
Grid-based models are a good example of learned positional encoding when the coordinates are bounded. 
They are defined by $d$-dimensional discrete grids of resolution $r \in \mathbb{N}$,
and store a vector of parameters at each discrete location in $[0, r-1]^d$. 
Grid encoders use the input coordinate to sample parameter vectors from their grid.
A range of different methods for representing the grids, 
sampling the grids, or applying various kernel functions to the resulting parameter vectors has been developed in recent years \cite{takikawa2021neural,zhao2024grounding}.
A known limitation of simple grids, which we highlight in the experimental section,
is that increasing their resolutions is almost the only way to increase their performance.
That was the motivation for several works designing multi-resolution grids, hash grids, etc.,
to capture broader frequency bands of the input coordinates \cite{muller2022instant}.

In this paper, we propose a simple method, Positional Encoding Projected Sampling (PEPS). 
Existing methods consider the result of absolute positional encoding as a high-dimensional vector that is either fed to the network or used as a multiplicative coefficient \cite{Fujieda23local}. 
PEPS instead considers it as a \textit{sequence of points} and uses each of those points to sample from an encoder. 
This formulation enables a basic grid representation to directly learn high-frequency signals, overcoming the need for high-resolution grids.
We propose an analysis of the motion of these points with respect to the multiplicative factors.
We show that their motion is composed of unique closed curves when sampled in equally space coordinates and contains useful geometric information. 
That is why PEPS applies \textit{learned positional encoding}
to each of the points resulting from \textit{absolute positional encoding}.
Moreover, in our experiments, PEPS uses fewer parameters than all existing methods while achieving better or similar accuracy.

Furthermore, PEPS comprises several components, such as the frequencies used, the encoders employed, and how multiple samples are aggregated. We propose using the power spectra of natural images to design an efficient aggregation function for PEPS.
Power Spectral Density (PSD) characterizes the power distribution of a signal over its constituent frequency components and serves as a useful metric for analyzing natural processes and learning capabilities. Remarkably, many real-world signals exhibit a PSD that follows an inverse relationship with frequency—a pattern observed across diverse phenomena ranging from natural images and musical melodies to biological system responses and even the electromagnetic emissions of massive quasars scattered across the distant universe \cite{keshner82,field1987relations,Szendro01012001}. We leverage this ubiquitous inverse relationship to design Pink-PEPS, a particular version of PEPS in which the resulting points are aggregated using a latent-dimension allocation inversely proportional to their frequency. For example, we demonstrate that the PSD of the Kodak image dataset \cite{kodak2022kodak} follows this inverse relationship, and our experimental section shows that Pink-PEPS is the best method for representing this dataset.

Finally, the experimental section focuses on three distinct applications.
The first application is an implicit image representation problem, a well studied problem consisting of storing an image into a neural model.
Using the Kodak dataset, we show that the PEPS methods lead to new highs in term of quality, error and also distance to the power spectra of the original images. 
In addition, we show that PEPS methods decrease performance by only 1\% compared to state-of-the-art grid methods while using 25\% fewer parameters.
The second application is 2D texture-set compression, where INRs are used to compress texture sets.
We compare PEPS against state-of-the-art models on texture compression and 
show that it improves on almost all texture types. 
Moreover, it reduces model size by more than 25\% while maintaining accuracy and rendering on par with the state of the art.
Third, for the 3D signed distance function problem,
the PEPS counterparts achieve the highest accuracy across all existing methods and strictly improve rendering on the hardest instances.
Specific contributions of this paper can be summarized as follows:  
\begin{itemize}
    \item A simple method named Positional Encoding Projected Sampling (PEPS) that applies a learned positional encoding to the output of absolute positional encoding.
    \item An extension named Pink-PEPS, which is faster and well suited for representing signals with inverse power spectral density. 
    \item Applications of PEPS to image representation and texture compression that strictly improve rendering quality over state-of-the-art models.
    \item Application of the PEPS method to signed distance function representation that achieves the highest accuracy in IoU and strictly improves performance on the hardest instances over state-of-the-art models.
\end{itemize}

The rest of the paper is organized as follows.
We provide background and motivation for our study in the next section, then present the theoretical framework for the PEPS method. We propose two example implementations: Grid-PEPS for wrapping grid encoders, and Pink-PEPS for noise reduction. Then, for each application, we provide a comparison against state-of-the-art methods and conclude.

\section{Positional Encoding and INRs}
The transformer architecture introduced by \cite{vaswani2017attention}
represented a paradigm shift, mainly due to the self-attention mechanism.
However, this architecture was unable to capture the positional information because of 
its permutation invariance.
That was the main motivation for the use of positional encoding (PE).
In their version, the one-dimensional position index $x$ is embedded into a high dimensional
space by applying sinusoidal functions to the position.
Let $\phi_i$ be the $i$th frequency targeted by the PE, the definition can be written:
\begin{align}
    \text{PE}(x,2i)&= \sin{x \phi_i}\\
    \text{PE}(x,2i+1)&= \cos{x \phi_i}
\end{align}
This definition, known as \textit{absolute positional encoding} (APE), 
was successfully incorporated in the learning process by
addition or multiplication to other embeddings 
or simply directly fed to neural networks.

Definitions for the $\phi$ values have been the focus of many research,
and various definitions, depending on the application, have been proposed.
In the context of LLMs, the very well known definition
$\phi_i=\frac{1}{10000^{i/d}}$,
with $d$ the number of dimensions of the positional encoding,
has been the reference since its definition.
In the context of neural fields and INRs, 
$\phi_i=2^i\pi$ 
has established dominance and is used in texture compression, NeRF,
and various higher dimensional problems \cite{mildenhall2021nerf}.

\paragraph{Learned positional encoding}
are methods that 
associate to each possible coordinate
a learned vector of parameters. 
It was stated in \cite{vaswani2017attention} that they used 
learned positional encoding analogously to \cite{gehring2017convolutional},
and results were identical.
Indeed, a drawback of learned positional encoding in the context of LLMs 
is their inability to represent a position unseen during training. 

Yet, in the INRs field, learned positional encoding 
has proven its efficacy compared to other methods,
and grid-based methods became the standard option \cite{zhao2024grounding}.
Grid-based methods uses grids, multi-grids, 
multi-resolution, or even multi-resolution-hash grids in their encoders.
They are a type of learned positional encoding 
as each coordinate will results in a learned parameter vector.
They are defined by three main components. 
The first one and most important one is the grid representation. 
A basic grid of resolution $r$ and dimensions $d$
stores $k$ dimensional feature vectors in each cell and can be represented by a 
tensor of shape $(r_1,\dots,r_d,k)$.
Many improvements of the grid representation have been proposed to reduce the memory consumption
by using hash-maps, quantized, or block compressed representation \cite{muller2022instant,nvidNTC23}.
The second component is the grid feature vector extraction, 
that given a coordinate, extracts a set of feature vectors from the grid.
For example, it is usual to extract the neighboring feature vectors of a given input coordinate.
Finally, the last component is the aggregation function that uses the 
input coordinate and the extracted feature vectors to output the final
feature vector. 
This last operation is often done by multi-linear interpolation of the neighboring feature vectors. 
Recent and state of the art results are made using a combination of those technics, 
combined with advanced quantization \cite{ubi24,laurent2025hardware,shi2025neural}.

\begin{figure}[ht]
    \centering
    \includegraphics[width=0.75\linewidth]{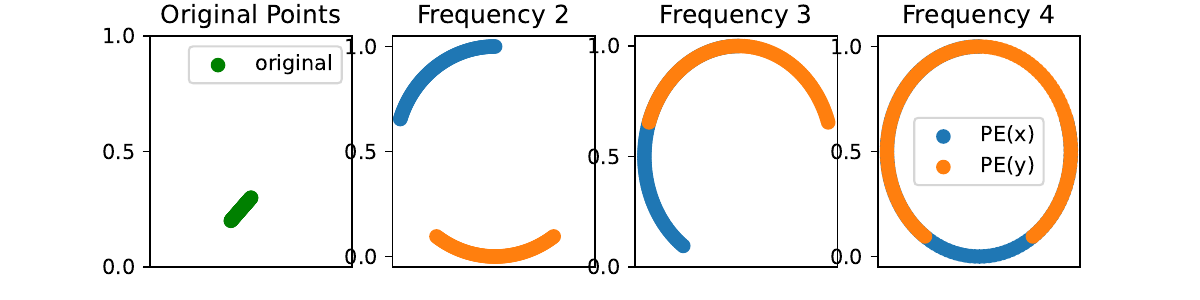}
    \caption{Rotation of $x$ and $y$ axis at each frequency. 
    On the first plot, the original line of points.
    Then on each frequency $i$ plot, the coordinates are $PE(x)=(\text{PE}(x,2i),\text{PE}(x,2i+1))$ 
    and normalized.}
    \label{fig:rotation}
\end{figure}

\paragraph{Advantage of APE}
The definition of the \textit{absolute positional encoding}
offers the advantage of encoding relative 1D distances.
Indeed, given a position $x$ and a position $x+k$,
the absolute positional encoding of these points is:
\begin{align}
    \text{PE}(x+k,2i)&= \text{PE}(x,2i)\cos{k\phi_i} + \text{PE}(x,2i+1)\sin{k\phi_i},\\
    \text{PE}(x+k,2i+1)&= \text{PE}(x,2i+1)\cos{k\phi_i} - \text{PE}(x,2i)\sin{k\phi_i}
\end{align}
This linear relationship between any two points at similar distance 
is one the of reason why this absolute definition allows to learn information
between tokens independently of their global position.
Moreover, this relationship is known to be a rotation using the following the matrix:
\begin{align}
\begin{bmatrix}
    \text{PE}(x+k,2i+1) \\
    \text{PE}(x+k,2i) 
\end{bmatrix}
=
\begin{bmatrix}
    \cos{k\phi_i} & -\sin{k\phi_i} \\
    \sin{k\phi_i} & \cos{k\phi_i}   
\end{bmatrix}
\begin{bmatrix}
    \text{PE}(x,2i+1) \\
    \text{PE}(x,2i) 
\end{bmatrix}
\end{align}
This rotation is \textit{axis wise}.
This implies that applying APE to higher dimensional coordinates,
such as 2D, results in each axis having this 
rotation relationship as it can be seen on Figure \ref{fig:rotation}.

As stated above, when considering both sinusoidal functions, 
the transformation is a rotation coordinate wise.
This implies that the positional encoding directly loss the link between the coordinates
of multi-dimensional points.
That is why for 2D and 3D problems such as chemical analysis, 
dedicated version are often designed to incorporate various geometric information 
\cite{wang2023geometric}.

\paragraph{Lissajous curves}
Even if the rotations of each axis is independent,
linear and simple, 
the multi-dimensional resulting spatial information is more complex.
Consider first the sinusoidal 2D case.
Given a point $(x,y)$, and a frequency $\phi_i$,
the generated point is $(\sin{\phi_i x},\sin{\phi_i y})$.
This is a particular case of Lissajous curve.
Lissajous curves \cite{lissajous1857memoire},
from the physicists Jules Antoine Lissajous,
are considered a particular case of complex harmonic motions and often used
in aerial search and polynomials interpolation \cite{steckenrider2024lissajous,erb2016bivariate2}.
An in depth study of Lissajous curves is out of the scope of this paper, yet more details are provided in appendix.
The main known property of those curves is the following.

\begin{proposition}\label{prop:LissajousUnique}
    For any two points $(x,y)$ and $(x',y')$, 
    their associated Lissajous curve is the same if and only if $(x,y)=(x',y')$.
\end{proposition}
The proof is given in appendix.

\section{Positional Encoding Projected Sampling}
We propose the Positional Encoding Projected Sampling (PEPS) method.
The previous section showed that the motion of points
under absolute positional encoding is geometrically meaningful and results in a unique trajectory for each point. 
Indeed, the 3D curve resulting from positional encoding is unique for each point and lies within the range [-1, 1].
The PEPS method projects the original point multiple times onto the Lissajous curve (these are the points of interest) and samples from an encoder at each of those points.
The objective is to use the uniqueness of the Lissajous curve to increase the information extracted by the learned positional encoding.

Consider INRs defined by an encoder (positional encoding, grid, etc.) 
and a subsequent model (MLP, etc.). 
PEPS can be seen as a wrapper around the encoder. 
It is a parametric method that projects the input coordinate into a series of points.
Then, each of these new points is fed to the encoder;
the results are aggregated and finally fed to the model.
These four steps (Project, Encode, Aggregate, and Model) are explicitly given in the following definition.

\paragraph{Definition}
Let $x \in \mathbb{R}^d$ be the $d$-dimensional input coordinate vector.
Let $L \in \mathbb{N}$ be the number of frequencies used in the absolute positional encoding.
Let the positional encoding coordinate points be defined by:
\begin{align}
    S_{i}(x) &= \frac{1+\sin{x \phi_i}}{2}\\
    C_{i}(x) &= \frac{1+\cos{x \phi_i}}{2}
\end{align}
for $i = 1,\dots,L$, where $\phi_i$ denote the frequency coefficients.
Let $P^x = (x, S_1,\dots,S_L,C_1,\dots,C_L)$ 
be the list of \textit{points of interest} of $x$.
$S_{i}$ is used instead of $S_{i}(x)$ when there is no ambiguity.
The size of $P^x$ is $2L+1$.
Let encoders $E = (E_1,\dots,E_{2L+1})$
be the list of encoders, and $M$ the model.
The PEPS method is defined by:
\begin{align}
    M\big(A(E_1(P^x_1),\dots,E_{2L+1}(P^x_{2L+1})), \delta\big),
\end{align}
where $A$ is the aggregation function and $\delta$ denotes the other inputs to the model.

\paragraph{Example: Grid-PEPS}
Let $G$ be a grid encoder, where each latent vector has dimension $k$.
Consider the case where $E_i = G$ for all $i$,
which implies that the grid is shared by all encoders.
Given a coordinate $x$, 
the Grid-PEPS method generates the points $P^x$, 
samples the grid $G$ at each element $P^x_i$ to 
obtain the latent vector $l^x_i$, and aggregates these results.
Figure~\ref{fig:gridpecs} shows the Grid-PEPS method,
using concatenation for aggregation, 
applied to neural texture compression.
Pseudocode is given in Algorithm~\ref{alg:pinkpecs}. 
The additional pink component is detailed in the next section.

The aggregation function $A$ plays an important role, 
analogous to techniques for embedding positional encoding into neural networks.
We show that the aggregation function can drastically improve results, incorporate task-specific knowledge, and affect runtime. 
We studied several aggregation functions (concatenation, additive learning, frequency based-additive learning, sub-vector extraction etc.).
The goal is to trade off accuracy against time and memory by controlling the number of features given to the MLP, as these methods are often coupled with very small MLPs (few layers with few neurons).
Unless otherwise specified, we use basic concatenation. 
Given that the dimension of each $l^x_i$ is $d$,
the final dimension of the extracted features is 
$(2L+1)d$.
As in absolute positional encoding, the value $L$ 
controls the total number of dimensions. 
The next section presents the Pink aggregator, 
an aggregation function that extracts smaller latent vectors.

\paragraph{Relationship with existing works}
If the encoders $E_i$ are identity functions, then the points are directly fed to the MLP, and this method is analogous to MLPs with absolute positional encoding. 
Hence, the proposed method can be seen as a generalization of APE. 
In the context of the combination of grids and APE, Fujieda et al. \cite{Fujieda23local} proposed the Local Positional Encoding (LPE) method.
The LPE method extracts the feature vector $l^x_0$ from the grid using the initial coordinate $x$, and returns the product $LocalPE(x)\odot l^x_0$, where $\odot$ is the Hadamard product, and $LocalPE(x)$ is the positional encoding of $x$ with respect to the local cell of the grid. 
This differs from the proposed method, which uses the different points resulting from the positional encoding to sample the grid multiple times.
In the context of hash grids, the input coordinate is transformed by the hash function to be restricted to the interval $[0,H]$, where $H$ is the hash-table size. 
The idea of using a transformation of the input coordinate before extracting information is related. However, these are two complementary methods, as the hash-map approach only deals with a single coordinate. 
Finally, multi-resolution grids use several grids of different resolutions, often in a pyramid setting.
They sample each of these grids with the input coordinate and aggregate the results. 
One possible comparison is that PEPS performs multi-frequency grids (encodings) in the sense that, instead of changing the resolution between grids to capture higher-frequency details, PEPS directly uses the frequency itself. 
We show in the experimental section that the proposed PEPS extension can be used on par with multi-resolution grids, hash grids, or any encoder, as it can be seen as a wrapper for these, and always leads to improvements.

\section{Pink Aggregator for Natural Processes (Pink-PEPS)}
The previous section defined the PEPS method.
In this section, we show how to design an aggregation function for PEPS 
to control the number of dimensions and the energy allocated to each frequency.

\begin{figure}
    \centering
    \includegraphics[width=.49\linewidth]{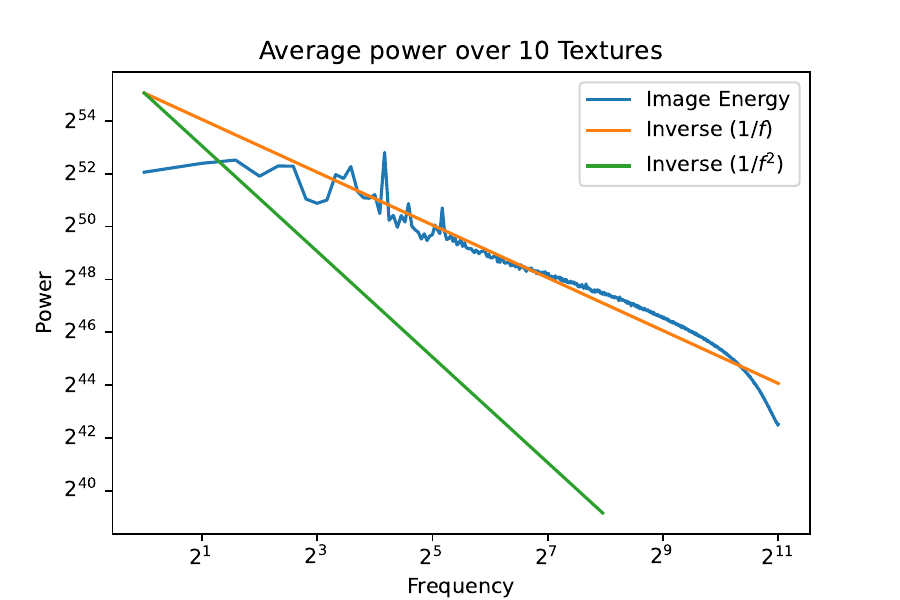}
    \includegraphics[width=.49\linewidth]{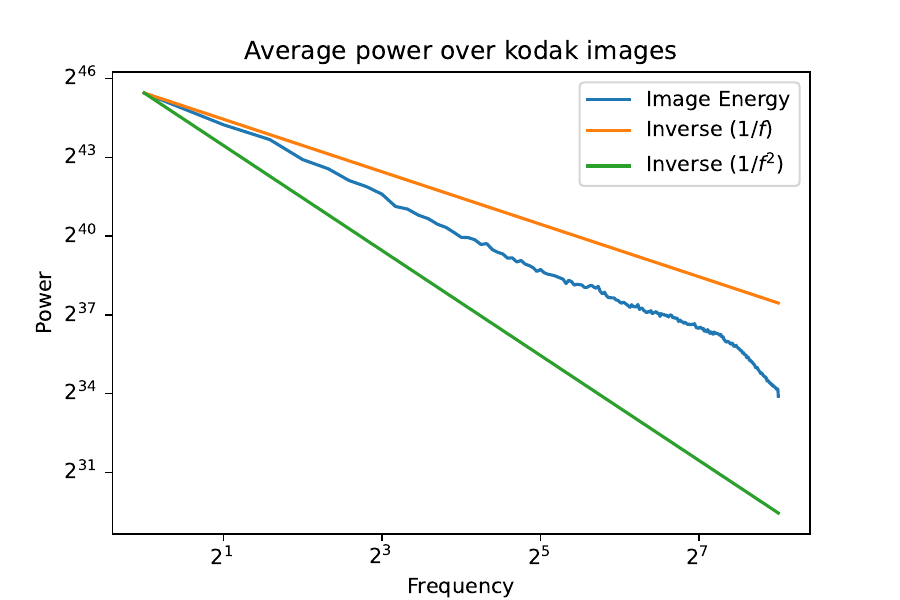}
    \caption{(left) Power spectra of 10 textures. (right) Power spectra of the Kodak dataset.
    As we can see, both are following a $\frac{1}{f^\alpha}$ law \cite{van1996modelling}.}
    \label{img:energy}
\end{figure}

\paragraph{Pink noise}
Fourier analysis has long been the backbone for analyzing signals, decomposing them into their frequency components.
One powerful metric based on Fourier analysis is the power spectral density (PSD).
From natural images and the emissions of massive quasars scattered across the universe to the responses of biological systems and the melodies of music, many real-world signals have a PSD that follows an inverse relationship with frequency \cite{keshner82,field1987relations,Szendro01012001}. 
These laws are usually described as $1/f^\alpha$ laws.
When $\alpha=0$, the power is the same across frequencies; white noise is an example. 
When $\alpha=2$, the signal is dominated by low frequencies; Brownian motion is an example of such a process, where the next position is a small random step from the previous one.
In between, when $\alpha=1$, we have pink noise, often simply written \oof noise. 
It is a distribution of power that is inversely proportional to frequency.
Generating \oof noise has been a major topic of interest in fields ranging from music \cite{voss75,voss78,pachet2015generating,perez2018extending}
to digital image generation \cite{perlin85}. 
For example, Figure~\ref{img:energy}
shows the power spectra analysis of our texture set dataset and of the Kodak dataset \cite{kodak2022kodak}
indicating that they also follow a $1/f$ law. 
More precisely, the texture set exhibits an initial plateau and then follows a \oof law,
whereas the Kodak natural images follow a $1/f^\alpha$ law with $1 \le \alpha \le 2$.
Given the potential benefits of generating \oof signals, several methods have been proposed  \cite{kasdin95,keshner82}.
A notably interesting algorithm is the Voss algorithm, which was developed by the physicist Richard Voss and then later published by Martin Gardner \cite{gardner78}.
These algorithms are often based on exponentially reducing the probability or rate of modification of higher-frequency components.

For PEPS, aggregating by directly concatenating the latent vectors resulting from each frequency gives the same potential power to each frequency. 
Therefore, we design an aggregation function to control the power allocated to each frequency sampled by the PEPS method. 
Intuitively, given two frequencies $f$ and $2f$, the inverse (\oof) law implies that the power at frequency $2f$ should be approximately half of that at $f$. 
We propose the Pink aggregator for PEPS (Pink-PEPS).

\paragraph{Pink-PEPS}
The Pink aggregator for PEPS is a resource-allocation strategy for the sampled latent vectors that takes into account the power spectra of natural processes such as images, reduces the method’s dimensionality, and ensures uniform gradient flow across the grid. 
More precisely, it is an aggregation function that extracts a shifted subvector from each latent vector produced by the encoders used by PEPS and concatenates them.

Given a vector $v \in \mathbb{R}^d$,
we denote by $v_{i:j}$ the circular sub-vector of $v$ containing elements 
$(v_{k_i}, v_{k_{i+1}}, \dots, v_{k_j})$ with $k_i=i\mod d$.
Let the sequence $a_0=0$, $a_n= \max(1, \left\lfloor \frac{d}{f_n}  \right\rfloor)$ for any strictly positive integer $n$,
denote the allocation of latent dimensions for each frequency $n$.
For example, using the Fourier encoding $\phi_i=2^i\pi$, 
the frequency if given by $f_i=\frac{\phi_i}{2\pi}=2^i$.
This implies that $a_n=\max(1, \left\lfloor \frac{d}{2^n}  \right\rfloor)$.

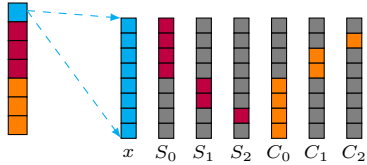
\begin{figure}
    \centering
    \begin{tikzpicture}
    
    \tikzmath{\s = 0.25; } ;
    \tikzmath{\c = 5.0; \v =2.60; } ;
    \draw[draw=black, fill=cyan, thin, solid] ({\c},{\v}) rectangle ({\c+\s},{\v-\s});
    \foreach \z in {1,2,3}
        {
	\draw[draw=black, fill=purple, thin, solid] ({\c},{\v-\s*\z}) rectangle ({\c+\s},{\v-\s*(\z+1)}) ;
    }
    \foreach \z in {4,5,6}
        {
	\draw[draw=black, fill=orange, thin, solid] ({\c},{\v-\s*\z}) rectangle ({\c+\s},{\v-\s*(\z+1)}) ;
    }

    \tikzmath{\z = 3; };
    \draw[draw=cyan, -latex, thin, dashed] ({\c+\s},{\v-0.1}) -- ({\c+1.5},{\v-0.2*1});
    \draw[draw=cyan, -latex, thin, dashed] ({\c+\s},{\v-0.1}) -- ({\c+1.5},{\v-0.2*9});
    
    \tikzmath{\s = 0.2; } ;
    \tikzmath{\c = 6.50; \v =2.60; } ;
    \foreach \z in {1,2,3,4,5,6,7,8}
        {
	\draw[draw=black, fill=cyan, thin, solid] ({\c},{\v-\s*\z}) rectangle ({\c+\s},{\v-\s*(\z+1)}) ;
    }

    \tikzmath{\c = 7; \v =2.60; } ;
    \foreach \z in {1,2,3,4}
        {
	\draw[draw=black, fill=purple, thin, solid] ({\c},{\v-\s*\z}) rectangle ({\c+\s},{\v-\s*(\z+1)}) ;
    }
    \foreach \z in {5,6,7,8}
        {
	\draw[draw=black, fill=gray, thin, solid] ({\c},{\v-\s*\z}) rectangle ({\c+\s},{\v-\s*(\z+1)}) ;
    }

    \tikzmath{\c = 7.5; \v =2.60; } ;
    \foreach \z in {5,6}
        {
	\draw[draw=black, fill=purple, thin, solid] ({\c},{\v-\s*\z}) rectangle ({\c+\s},{\v-\s*(\z+1)}) ;
    }
    \foreach \z in {1,2,3,4,7,8}
        {
	\draw[draw=black, fill=gray, thin, solid] ({\c},{\v-\s*\z}) rectangle ({\c+\s},{\v-\s*(\z+1)}) ;
    }

    \tikzmath{\c = 8; \v =2.60; } ;
    \foreach \z in {7}
        {
	\draw[draw=black, fill=purple, thin, solid] ({\c},{\v-\s*\z}) rectangle ({\c+\s},{\v-\s*(\z+1)}) ;
    }
    \foreach \z in {1,2,3,4,5,6,8}
        {
	\draw[draw=black, fill=gray, thin, solid] ({\c},{\v-\s*\z}) rectangle ({\c+\s},{\v-\s*(\z+1)}) ;
    }

    \tikzmath{\c = 8.5; \v =2.60; } ;
    \foreach \z in {5,6,7,8}
        {
	\draw[draw=black, fill=orange, thin, solid] ({\c},{\v-\s*\z}) rectangle ({\c+\s},{\v-\s*(\z+1)}) ;
    }
    \foreach \z in {1,2,3,4}
        {
	\draw[draw=black, fill=gray, thin, solid] ({\c},{\v-\s*\z}) rectangle ({\c+\s},{\v-\s*(\z+1)}) ;
    }

    \tikzmath{\c = 9; \v =2.60; } ;
    \foreach \z in {3,4}
        {
	\draw[draw=black, fill=orange, thin, solid] ({\c},{\v-\s*\z}) rectangle ({\c+\s},{\v-\s*(\z+1)}) ;
    }
    \foreach \z in {1,2,5,6,7,8}
        {
	\draw[draw=black, fill=gray, thin, solid] ({\c},{\v-\s*\z}) rectangle ({\c+\s},{\v-\s*(\z+1)}) ;
    }

    \tikzmath{\c = 9.5; \v =2.60; } ;
    \foreach \z in {2}
        {
	\draw[draw=black, fill=orange, thin, solid] ({\c},{\v-\s*\z}) rectangle ({\c+\s},{\v-\s*(\z+1)}) ;
    }
    \foreach \z in {1,3,4,5,6,7,8}
        {
	\draw[draw=black, fill=gray, thin, solid] ({\c},{\v-\s*\z}) rectangle ({\c+\s},{\v-\s*(\z+1)}) ;
    }
    
    \node at ({6.6},{\v-\s*10})  {\scriptsize$x$};
    \node at ({7.1},{\v-\s*10})  {\scriptsize$S_0$};
    \node at ({7.6},{\v-\s*10})  {\scriptsize$S_1$};
    \node at ({8.1},{\v-\s*10})  {\scriptsize$S_2$};
    \node at ({8.6},{\v-\s*10})  {\scriptsize$C_0$};
    \node at ({9.1},{\v-\s*10})  {\scriptsize$C_1$};
    \node at ({9.6},{\v-\s*10})  {\scriptsize$C_2$};
    
\end{tikzpicture}
    \caption{Pink aggregation applied to a latent dimension ($d$) of 8, and using the Fourier encoding ($\phi_i=2^i\pi$).
    The simple concatenation would directly concatenate the 8*7=56 values (i.e., the blue, purple, orange, and gray ones), while the Pink-PEPS only considers 8+2*4+2*2+2*1=22 values that are inversely spread given their frequency (i.e., the blue, purple, and orange ones).}
    \label{fig:Pinkpecs}
\end{figure}

Let the series $G_n=\sum_{i=0}^n a_i$ denotes the sum of already 
allocated dimensions for frequencies 1 to $n$.
The Pink aggregator extracts from each vector $l^{C_i}$ (resp.  $l^{S_i}$) 
the sub-vector $l^{C_i}_{G_{i-1}:G_{i}}$ (resp.  $l^{S_i}_{-G_{i}:-G_{i-1}}$).
An example is given in Figure~\ref{fig:Pinkpecs}, and 
a possible pseudo-code is given by Algorithm~\ref{alg:pinkpecs}.

Intuitively, for each higher frequency,
the Pink-PEPS method restricts the dimensionality of the sampled latent vector so that it is inversely proportional to the frequency. 
For example, using the Fourier positional encoding, the latent dimension is divided by two each time and shifted. 
The intuition behind the shift, using the series $G_n$, is to ensure that, when sharing a grid or any learned PE, the whole latent vector receives gradients, rather than only the first few dimensions. 
Note that the complete vectors (e.g., $l^{C_i}$) do not need to be computed, as only subparts are required. 
This is the particular point that makes our implementation of Pink-PEPS faster.
Note that an $\alpha$ term can be added to the definition of $\frac{d}{f_n^\alpha}$ for $a_n$.
When $\alpha=0$, this is the usual PEPS; when $\alpha=1$, this is Pink-PEPS; 
and when $\alpha=2$, this gives Brownian-PEPS. 
This definition generalizes the PEPS method.

\begin{algorithm}
\caption{Full {\color{gray}Grid}-{\color{magenta}Pink}-PEPS algorithm}\label{alg:pinkpecs}
\begin{algorithmic}
\Require $x \in \mathbb{R}^d, {\color{gray}E}:$ grid encoder, $L \in \mathbb{N}$
\State $g \gets {\color{gray}E}(x)$
\For{$i = 1, \dots, L$}
\State $l^{S_i} = {\color{gray}E}(\sin{x \phi_i})$
\State $l^{C_i} = {\color{gray}E}(\cos{x \phi_i})$
\State $g.\text{append}(l^{S_i}{\color{magenta}_{-G_{i}:-G_{i-1}}})$
\State $g.\text{append}(l^{C_i}{\color{magenta}_{G_{i-1}:G_{i}}})$
\EndFor
\State \Return $g$
\end{algorithmic}
\end{algorithm}

\section{Application: Implicit Image Representation}
We start the experimental section with the implicit image representation problem.
In implicit image representation, given a coordinate $x\in [0,1]^2$, 
the function $P(x)$ returns the RGB values $(r,g,b) \in \mathbb{R}^{3}$ at location (x).
Implicit neural representation is the set of methods using neural networks for representing function $P$. 
This section is split in two parts.
First we propose to analyze the learning capability of methods based on grids, 
and shows how the PEPS method is able to improve their overall capabilities. 
Then, we compare using the Kodak images dataset \cite{kodak2022kodak}
to prove the efficacy of the proposed method in the usual settings.

\paragraph{Current bottlenecks for grid methods}
Before going into the large comparison against the state of the art methods,
we propose to first analyze the parameters impact of the grid methods and PEPS.
We use bilinear Interpolation grids (BI grid) 
which are grids using the bilinear interpolation of the neighbors latent vectors.
Local Positional Encoding (LPE) is a wrapper around a BI Grid and multiplies the 
resulting latent vector by the positional encoding of the point with respect to the cell \cite{Fujieda23local}.
Grid-PEPS is a PEPS wrapper around the same BI Grid which is shared across frequencies and using three frequencies $(\phi_1,\phi_2,\phi_3)$.
The $\phi_i=2^i\pi$ definition is used in all experimental sections.
The models are composed of the encoder followed by an MLP of three hidden layers of 64 neurons for both experiments.
The learning rate is 0.01. More details are given in the technical appendix.
The first experiment is on learning  4K images  (roughly $10^8$ values) with models having
between $10^4$ to $10^6$ parameters (grid resolutions between 16 to 128, feature vector dimensions between 8 to 64).
For each image, a model is trained using the same number of steps, 
and the results are with respect to the average of all the runs of all images. 
All implementations have been made in pytorch \cite{paszke2017automatic}
and are shared between methods to avoid artifact due to different initialization, etc.
The loss is the $L_1$ loss.
The metric used in this first study is the Peak signal-to-noise ratio (PSNR).
PSNR being based on the log of the mean square error, 
it is a good learning capability metric for neural networks. 
Positional encoding is kept out of these results as it does not have a parameter other than $L$,
and is strictly dominated by several order of magnitude when using small neural networks.
Using larger neural network lightly improves the results, yet not as much as grid-based methods and
drastically increase the running time.
%
\begin{figure}[h]
  \centering
  \includegraphics[width=0.6\linewidth]{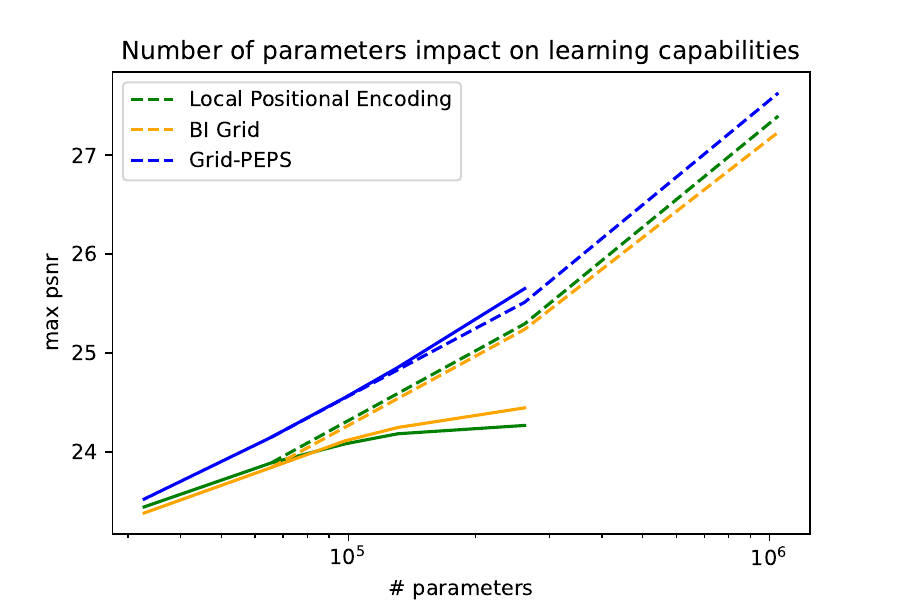}
  \caption{Impact of the number of parameters on the image learning capability (PSNR$\uparrow$).
    Dashed lines are grid resolution modifications.
    Regular lines are feature dimension modifications.}
    \label{fig:paramImpact}
\end{figure}
%
In such setting of extreme compression,
a hypothesis would be that additional parameters should always improve the model.
Yet, Figure~\ref{fig:paramImpact} shows that increasing the parameters doesn't always help the learning model.
Even in this setting,
methods such as grid and LPE fail to improve when given 
higher dimensional latent vectors. 
In contrary, the Grid-PEPS method, which is based on the BI grid, 
is able to linearly use the additional parameters to improve its fitting.
It is able to reach new highs in PSNR with low resolution grids.
In addition, the recent paper on NTBC \cite{laurent2025hardware} 
explain that they had to shift by half a texel the odd textures, etc. 
This kind of issue is by design corrected by the PEPS methods.
Finally, PEPS remains the best method, whatever the parameters configuration.

\begin{table*}[!ht]
\centering
\scriptsize
\caption{Kodak dataset of \textit{low} resolution (768,512) images. All methods have the same number of parameters, except LPE having 6 percent less parameters, the bottom row having 25 percent less parameters, and PE which use a larger MLP.}
\label{tab:ntckodak}
\begin{tabular}{|l|c|c|c|c|c|}
\hline
Method & PSNR ($\uparrow$)& FLIP ($\downarrow$)& LPIPS ($\downarrow$)& LSD ($\downarrow$)&  SSIM ($\uparrow$)\\ \hhline{|=|=|=|=|=|=|}
PE  & 39.91 & 3.71e-02 & 1.47e-02 & 4.46e-02 & 0.960  \\ \hline
LPE -6$\%$ & 45.06 & \textit{1.84e-02} & 1.62e-03 & 7.49e-03 & 0.992 \\ \hline
NTC$_{N}$  & 44.87 & 1.96e-02 & 2.23e-03 & 9.74e-03 & 0.992 \\ \hline
Grid  & 45.30 & \textbf{1.83e-02} & 1.37e-03 & 1.24e-02 & 0.992 \\ \hhline{|=|=|=|=|=|=|}
G-PEPS  & 47.72 & 2.08e-02 & 1.30e-03 & \textit{4.19e-03} & \textit{0.993} \\ \hline
G-P-PEPS  & 47.83 & 2.24e-02 & \textbf{1.05e-03} & 4.41e-03 & \textit{0.993} \\ \hhline{|=|=|=|=|=|=|}
NTC$_{PEPS}$  & \textit{48.02} & 2.07e-02 & 1.43e-03 & \textbf{4.07e-03} & \textbf{0.994} \\ \hline
NTC$_{PinkPEPS}$  & \textbf{48.07} & 2.14e-02 & \textit{1.20e-03} & 4.22e-03 & \textbf{0.994} \\ \hhline{|=|=|=|=|=|=|}
G-P-PEPS -25$\%$  & 44.89 & 2.81e-02 & 3.12e-03 & 8.45e-03 & 0.987 \\ \hline
\end{tabular}
\end{table*}

\paragraph{Kodak dataset}
The final MLP having the same size for any final resolution 
might results in discrepancies depending on the resolution.
We propose to focus on a smaller resolution set, with smaller encoders
but using the same MLP size as before (three hidden of 64 neurons) to avoid its saturation and highlight the actual impact of the encoder. 
The dataset is the Kodak dataset having 24 images of resolution (768,512) \cite{kodak2022kodak}.
The NTC methods have two grids of respective size (192,128) and (96,64) 
and respective feature dimensions of 12 and 20 following the architecture from the neural texture compression paper \cite{nvidNTC23}.
Grid and G-PEPS and G-P-PEPS (Grid Pink PEPS) have the same grid of (196,128) and 17 features.
Those settings represents a factor of around 2.7 of compression. 
The PE (positional encoding) used 10 Fourier frequencies and a larger MLP of 3 layers of 300 neurons.
Four additional metrics are reported here.
The SSIM metric is a perception quality metric comparing the structural information of the images \cite{wang2004image}.
The FLIP error is the average of the FLIP error of an image \cite{andersson2020flip}.
The LPIPS error is a perception metric defined using an Alexnet \cite{zhang2018unreasonable}.
The LSD metric, also known as log-spectral distortion, is simply the distance between two log-spectra applied to the 2D Fourier representation of the image \cite{rabiner1993fundamentals,gray2003distance}. 
This measure is heavily used in signal processing, and mostly in speech applications. 
More discussion about those metrics can be found in appendix.
We use this spectral metric here to characterize the capability of a neural network to output an image having the same spectral property as the original one.
As we can see in Table \ref{tab:ntckodak}
the PEPS method (G-PEPS and NTC$_{PEPS}$) are drastically improving their non-PEPS counterparts.
Then, the PinkPEPS methods (G-P-PEPS, etc.) are clearly better in all metrics including reconstruction error and perceptual quality and spectral distance.
Our current explanation is that in all settings, the additional features brought by PEPS methods are informative,
but the small neural networks required for real-time rendering might struggle to use it. 
The PinkPEPS methods returning less values, especially less high frequency values, reduces the input noise and the total number of parameters.
Dual comparison scatters are provided in Figure~\ref{fig:scatterKodak}.
As we can see, not only the PEPS method is better in average, but it completely dominates its grid counterpart. 
Moreover, the version having 25\% less parameters seems comparable to the existing SOTA in PSNR, LSD, and SSIM.
The same result conclusion can be seen for the NTC versions too.
Moreover, even with a L2 loss, adaptive dual learning rates and different activation (full description in appendix), the PEPS remains dominant as shown by the fourth plot.

\begin{figure*}
    \includegraphics[width=0.95\linewidth]{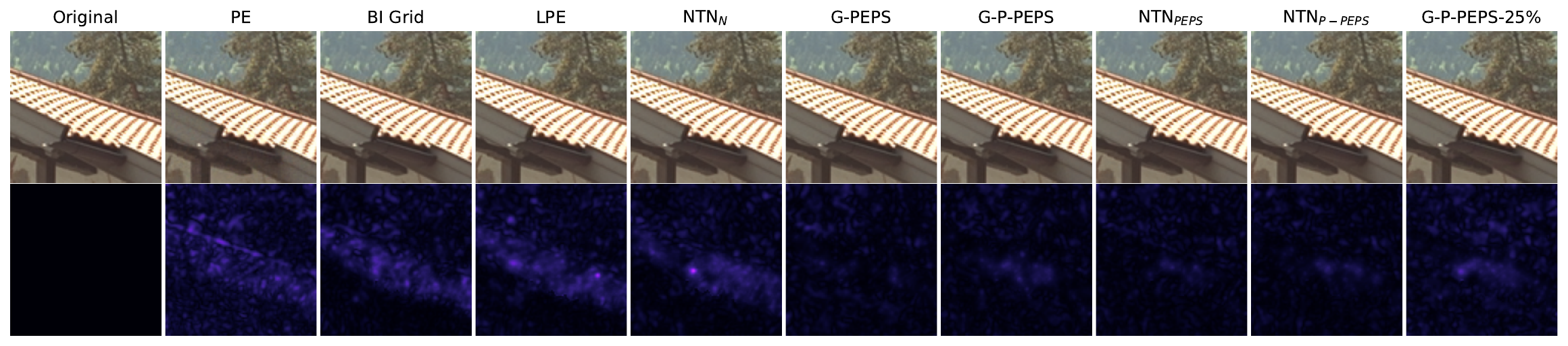}
    \caption{(Image) 100 pixels samples from the Kodak dataset. Bottom images are Flip error \cite{andersson2020flip}.}
    \label{fig:sampleKodak}
\end{figure*}

\begin{figure}
    \includegraphics[width=0.99\linewidth]{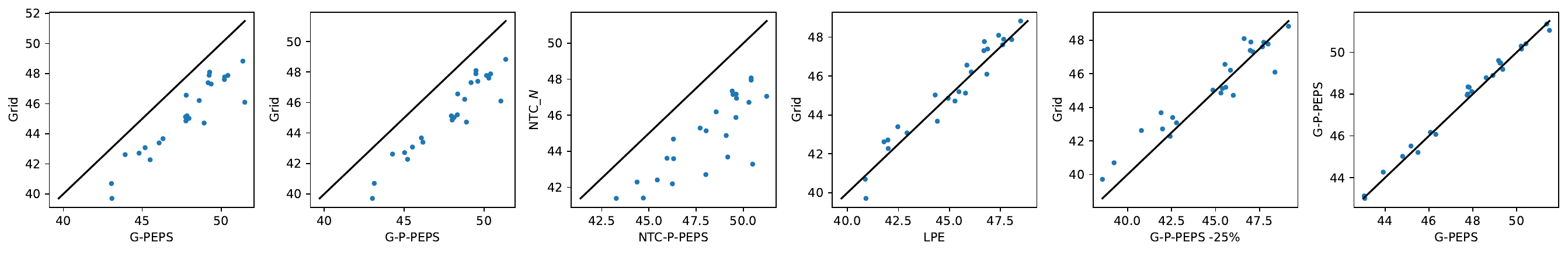}
    \caption{Dual scatters of PSNR results for the Kodak dataset.
    Each axis is representing the PSNR of a method. 
    If a point is on the side of a method, its result is better for this method.
    }
    \label{fig:scatterKodak}
\end{figure}

In addition, the result that the basic grid is better than the engineered method NTC$_{N}$
can be directly explained by Figure~\ref{fig:paramImpact}.
In this settings, 
trading the features of higher resolution grids to make high-dimensionality latent features in lower-resolution grids
is not working, and the NTC$_{N}$ lacks efficiency. 
That is the bottleneck we highlighted.
In contrary, as shown in Figure~\ref{fig:paramImpact}, 
the PEPS method is able to use the low-resolution grids with
high-dimensional latent feature vectors, which explains the good results of NTC$_{PEPS}$.
That is one of the main advantages of the PEPS method, the ability to use parameters almost equivalently from resolution of latent dimensional size.
Samples from the generated images are shown in Figure~\ref{fig:sampleKodak},
and in the additional files.
As we can see, the PEPS methods exhibit less error and are closer to the color palette from the original image.
The results of this section aimed at highlighting the impact of the PEPS method.
We also tested other loss functions (L1, L2), activation functions (Leaky Relu, Gelu, Silu), and adaptive dual learning rates. 
None of those modification led to change in the capability of the PEPS counterpart to perform better in term of PSNR and SSIM.
Some of results are shown in appendix.


\section{Application: Neural Texture Compression}
Textures and materials are the backbone of modern 
3D engines and physically based rendering that are used in applications such as games.
In those engines, many 4K textures must be loaded at the same time,
creating a memory bottleneck.
That is the reason why neural texture compression (NTC) has gained interest.
Compared to traditional image compression technics, the goal is to 
have random access to the values of all the textures of the set at some given pixel locations.
Indeed, the rendering mechanism usually sample the information of the texture locations hit by rays.
Recently, an emphasis has been put on texture representation and compression,
and existing works has shown that INRs can represents textures sets at different MIP level efficiently.
\cite{nvidNTC23,farhadzadeh2024neural}.
Preliminary research works have focused on the capability of positional encoding and 
grid-like representation to represent and efficiently compress textures and materials.
Recent works focused more on making the grid-like methods faster by
quantization-like technics either ad-hoc or using available tools such as 
block compression and cooperative vectors \cite{ubi24,shi2025neural,laurent2025hardware}.
These methods can be described using their multi-grids definition, 
fine tuned frequencies for positional encoding,
and quantization strategy.
From a learning capability point of view, 
even if their implementation is different,
they are all bounded by the learning capability of the grid itself.
Compared to image representation, in texture representation, 
given a coordinate $x\in [0,1]^2$, 
the function $T(x)$ returns the texture set information $t \in \mathbb{R}^{3k}$ 
at this location with $k$ the number of textures of the texture set.
The goal for a given texture set $S$ associated with function $T_S$,
is to train a network $N_S$ such that
$N_S(x)=T_S(x), \quad \forall x \in [0,1]^2$.
Additional goals are compression and real-time capabilities.

\begin{figure}[ht]
    \centering
    \includegraphics[width=1.\linewidth]{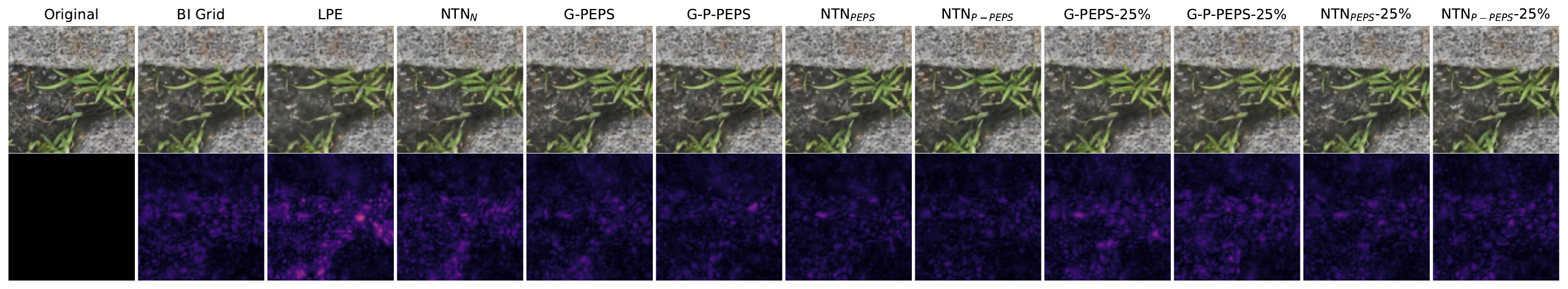}
    \caption{100 pixels samples from the 4K Paving Stone texture set. Bottom images are Flip error \cite{andersson2020flip}.}
    \label{fig:samplecard}
\end{figure}

\paragraph{Texture set compression}
For a fair comparison, 
we propose to align all the methods with the configuration given in \cite{nvidNTC23} named NTC$_{N}$.
For 4K texture sets,
the BI Grid, Grid-PEPS, and Grid-Pink-PEPS methods will have a single grid of resolution 1024 and feature vectors of dimension 17.
The LPE method will have a single grid of resolution 1024 and feature vectors of dimension 16.
The NTC methods have two grids of respective size 1024 and 512 and respective feature dimensions of 12 and 20.
In addition, the NTC methods have the \textit{Tiled positional encoding} which corresponds here to the 
last three frequencies of the positional encoding with respect to image size.
The median number of textures in texture sets is 5, so the average compression factor is 14 (ratio is 93\%).
The "-25\%" versions have the same grid sizes and respective latent-feature dimensions of 9 and 15 for the NTCs and 13 for the Grids,
We experimentally saw that the number of frequencies have a light impact on the PSNR, and almost none on the SSIM, 
yet three or four frequencies seems to be a sweet spot for the methods.
In this section, all PEPS methods uses four frequencies.
so the average compression factor is 18 (ratio is 95\%).
No quantization is involved as our goal is to analyze the impact of the PEPS methods,
and not the various quantization strategies.
For that reason, methods from \cite{ubi24,laurent2025hardware,shi2025neural}
as they rely on quantization technics of grids are omitted.
The rest of the configuration is the same as the previous experiment.
The model is learning the filtered textures directly using bilinear interpolation of the true pixels
during the training.
The texture sets used in this paper are based on the available sets from  \cite{fujieda2024neural,ubi24}
and additional sets, all available from https://polyhaven.com/. 
Except when clearly stated, the total number of texture sets is 18.
Full description of the dataset is given in appendix. 

\begin{table*}[ht]
\centering
\scriptsize
\caption{
For each texture type (AO, etc.) the value is the average PSNR over the dateset.
PSNR and SSIM columns are the average over all textures of all texture sets of the dataset.
NTC$_{N}$ is the method from \cite{nvidNTC23} without quantization.
LPE is the method from \cite{Fujieda23local}.
The three first line are SOTA methods. 
Then, the PEPS method is the middle are wrapped around the BI grid or using the same multi-resolution grid architecture as \cite{nvidNTC23}.
Finally, at the bottom, all those methods have have 25\% fewer parameters that the rest of the table that have the same overall number of parameters.
}
\label{tab:ntc}
\begin{tabular}{|l||c|c|c|c|c|c|c|c|c|c|c|}
\hline
Method  & PSNR & SSIM & AO & ARM & DIFF & Disp. & metal & normals & rough & specular  \\  \hhline{|=|=|=|=|=|=|=|=|=|=|=|}
LPE  & 40.21 & 0.95 & 42.16 & 41.18 & 36.38 & 50.34 & 46.83 & 34.99 & 39.05 & 44.76 \\ \hline
NTC$_{N}$  & 40.20 & 0.95 & 42.59 & 41.29 & 36.78 & 48.99 & 46.07 & 34.99 & 39.18 & 46.13 \\ \hline
BI grid  & 41.25 & 0.95 & 43.59 & 42.24 & 37.42 & \textbf{51.18} & 47.35 & 36.03 & 39.98 & 46.07 \\ \hhline{|=|=|=|=|=|=|=|=|=|=|=|}
Grid-PEPS4F  & 41.23 & 0.95 & 43.42 & 42.11 & 37.09 & \textit{50.81} & 49.90 & 35.90 & 39.67 & 47.74 \\ \hline
Grid-PinkPEPS4F  & 41.44 & 0.95 & 43.55 & 42.48 & 37.48 & 50.64 & 49.43 & 36.24 & 40.10 & 47.73 \\ \hline
NTC$_{PEPS}$  & \textit{41.79} & 0.95 & \textit{44.17} & \textit{42.73} & \textit{37.91} & 50.28 & \textit{50.47} & \textit{36.64} & \textit{40.23} & \textbf{48.56} \\ \hline
NTC$_{PinkPEPS}$  & \textbf{41.89} & 0.95 & \textbf{44.34} & \textbf{42.79} & \textbf{38.09} & 50.42 & \textbf{50.10} & \textbf{36.71} & \textbf{40.47} & \textit{48.24} \\ \hhline{|=|=|=|=|=|=|=|=|=|=|=|}
Grid-PEPS4F-25\%   & 39.86 & 0.93 & 42.19 & 40.83 & 35.86 & 49.15 & 48.63 & 34.58 & 38.50 & 45.41 \\ \hline
Grid-PinkPEPS4F-25\%   & 40.03 & 0.94 & 42.32 & 41.26 & 36.20 & 48.83 & 48.07 & 34.81 & 38.83 & 45.31 \\ \hline
NTC$_{PEPS}$-25\%  & 40.59 & 0.94 & 43.17 & 41.60 & 36.70 & 49.65 & 49.23 & 35.27 & 39.10 & 46.15 \\ \hline
NTC$_{PinkPEPS}$-25\%   & 40.56 & 0.94 & 43.18 & 41.65 & 36.74 & 49.61 & 47.90 & 35.30 & 39.19 & 46.12 \\ \hhline{|=|=|=|=|=|=|=|=|=|=|=|}
\end{tabular}
\end{table*}

Table~\ref{tab:ntc} contains the averaged results over our dataset.
As we can see, the Grid Pink PEPS method has better quality metrics
than both the BI grid and the NTC$_{N}$ methods containing several grids and positional encoding.
This is a clear result that shows the proposed method improves the learning capabilities
of grid-based methods.
For texture compression, the PinkPEPS improvement of gridPEPS is smaller, yet still present.
A reason could be that the PSD of the texture set starts with a plateau as shown in Figure~\ref{img:energy}.
The NTC$_{PEPS}$ and NTC$_{PinkPEPS}$ methods that use the architecture of NTC$_{N}$,
and Grid-PEPS or PinkPEPS instead of the original grids, show the best performances overall. 
In almost all categories or texture type, they are strictly better.
This confirms that both from a direct use (i.e., grid pink PEPS),
and used inside high-level architecture, the proposed PEPS methods 
exhibit better learning capabilities.
Finally, the appendix contains samples of the resulting textures.
They show that the PEPS counterparts almost always have better FLIP error.
Figure~\ref{fig:samplecard} is one of them.
In conclusion, in all three compression settings, extreme with bottleneck, 
low with Kodak dataset, and regular compressions here,
using the PEPS alternatives always improve the reconstruction error.

\paragraph{Runtime Analysis}
We evaluated the performance overhead of PEPS by implementing neural texture decompression using HIP \cite{hip_github} and running the code on a Radeon \texttrademark RX 9070 XT GPU. The implementation uses Wave Matrix Multiply Accumulate intrinsics for the MLP \cite{amd_rdna4_isa_2025}. The encoder and the MLP were fused into a single kernel. 
We compared the inference time of BI grid and Grid-PEPS, and Grid-PinkPEPS with the same grid configuration:
a $1024\times 1024$ grid, a 16 dimensional feature vector and three frequencies for both Grid-PEPS, and Grid-PinkPEPS. Generating a single three-channel $1024^2$ texture took 5.47ms, for Grid-PEPS, 4.32ms for BI grid, and 4.86ms for Grid-PinkPEPS. Grid-PEPS was 26.6$\%$ slower than BI grid, a penalty that stems from the additional arithmetic operations and memory access required for the the additional grid sampling. Because Grid-PinkPEPS creates smaller input feature vector, it performs fewer arithmetic operation and memory accesses than Grid-PEPS, its runtime of 4.86ms corresponds to a 12.5 $\%$ overhead relative to the BI grid. We also tested Grid-PinkPEPS with four frequencies which results in 4.99ms (15.5$\%$ overhead), which is still faster than Grid-PEPS. When the grid size is smaller which makes it more cache friendly, we expect the overhead to be smaller.

\section{Application: Signed Distance Function}
Signed Distance Functions (SDFs) are used to represent 3D shapes during rendering \cite{osher2003signed,vicini2022differentiable}.
SDFs give the closest distance to surfaces for any spatial location to represent a shape.
In addition to distance, SDFs give the information of the location of the point.
The SDF will be positive when the point of interest is in the exterior region, and 
negative when in the interior region.
Many works have been defined for representing SDF 
as the dense representation requires a high-resolution grid that usually consumes 
a large amount of memory \cite{saragadam2023wire}.

Analogously to the previous section, we propose to analyze the impact of the
introduction of the PEPS method using existing encoders of the state of the art.
We will be using Trilinear-interpolated grids (\textit{TI grid}).
Hash-grid methods, multi-grids, and multi-resolution-hash grids \cite{muller2022instant}
are also compared.
Finally, we propose their PEPS versions, \textit{Grid-PEPS} for grid, \textit{Hash-PEPS} for Hash, 
\textit{m-PEPS} for multi-grids PEPS,
\textit{m-hashPEPS} for multi resolution hash grids with PEPS.
As for the NTC section, the PEPS method uses the same encoder across three frequencies.
No additional parameters are required except for the first layer of the MLP 
as the input is high-dimensional. 
We follow the experimental protocol setting from \cite{Fujieda23local}.
The single grid methods have a resolution of 32 and stores 18 dimensional feature vectors in each grid location.
The hash grid method has a resolution of 64 and stores 18 dimensional feature vectors in a hash table of size $32^3$.
The multi-grid method has 3 grids of resolution [16,32,64]
and stores 2 dimensional feature vectors in each grid location.
The multi-hash grids are of resolutions [16,32,64,128] and use 2 dimensional feature vectors
and maximum hash table size of $2^{17}$.
PE is a positional encoding with 10 frequencies.
Finally, the MLP is composed of 3 layers of 64 neurons.
The loss of this section is the mean absolute percentage error (MAPE) loss. 
The four SDF instances are defined using $512^3$ values.
The compression factor is 227 (ratio is 99.6\%).
Rendered results and results with \textit{L1} loss 
are present in the appendix and supplementary materials.
Figure~\ref{fig:sampleArma} is one of those samples.

\begin{figure*}
    \includegraphics[width=1\linewidth]{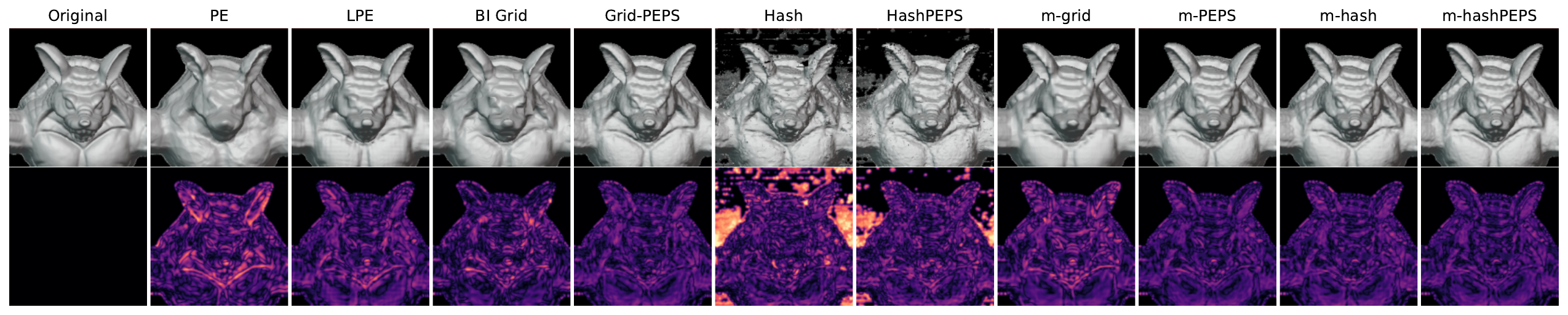}
    \caption{(SDF) 100 pixels samples from the Armadillo model. Bottom images are Flip error \cite{andersson2020flip}.}
    \label{fig:sampleArma}
\end{figure*}

\begin{table*}[ht]
\centering
\caption{(SDF) IoU ($\uparrow$) results for all the methods and our PEPS counterparts. Loss is MAPE.}
\label{tab:SDF}
\tiny
\begin{tabular}{|l|l|l|l|l|l|l|l|l|l|l|l|l|}
\hline
 Model & PE & LPE & TI Grid & Grid-PEPS & Hash & hashPEPS & M-Grid & M-PEPS & M-Hash & M-HashPEPS \\ \hhline{|=|=|=|=|=|=|=|=|=|=|=|}
 Lucy & 0.893 & 0.902 & 0.904 & \textit{0.909} & 0.640 & 0.876 & 0.902 & 0.908 & \textbf{0.910} & \textbf{0.910} \\ \hline 
 Pitted Stonefish & 0.316 & 0.389 & 0.410 & \textbf{0.470} & 0.339 & 0.428 & 0.413 & \textit{0.469} & 0.460 & \textbf{0.470}\\ \hline 
 Thai Statue & 0.918 & 0.925 & \textit{0.926} & \textbf{0.930} & 0.574 & 0.831 & 0.924 & \textbf{0.930} & \textbf{0.930} & \textbf{0.930}\\ \hline 
 Armadillo & 0.952 & \textit{0.956} & 0.955 & \textbf{0.957} & 0.637 & 0.788 & \textit{0.956} & \textbf{0.957} & \textbf{0.957} & \textbf{0.957}\\\hhline{|=|=|=|=|=|=|=|=|=|=|=|}
 
 Global & 0.770 & 0.793 & 0.799 & \textit{0.816} & 0.547 & 0.731 & 0.799 & \textit{0.816} & 0.814 & \textbf{0.817} \\  \hline 

\end{tabular}
\end{table*}

Table~\ref{tab:SDF} presents the overall results of the standard IoU metric for all the methods.
Compared to the image and NTC cases, the use of multi-grids is not improving IoU results for SDF,
but it is often improving the final rendering.
Again, in all experiments, the PEPS counterparts show better IoU and better quality rendering.
For the TI Grid, the introduction of PEPS improves the details of the render to a level of details
higher than the multi-resolution grids.
For the Hash, it is able to reduce the artifacts due to the hash table collision.
Finally, for both multi-resolution versions, the introduction of PEPS both improved the metrics 
and the quality of rendering.
Finally, the PEPS method seems to help unlock previously hard instances such as 
Pitted stonefish where the low resolution usually completely fails to render.
More precisely, Table~\ref{tab:SDFParamAppendix} shows the impact of increasing the number parameters of all the methods (except PE) for the hard instance Pitted Stonefish.
The goal of such a table is to show that the PEPS methods are able to achieve high reconstruction and that most of the other methods with 8 times more parameters are at the level of the grid-PEPS with 8 times less parameters. 
More results on that are provided in the appendix.

\begin{table*}[ht]
\centering
\caption{(SDF) IoU ($\uparrow$) results for all the methods using L1 loss. The bold line uses 8 times more parameters in the encoders (grid of resolution 64 instead of 32).}
\label{tab:SDFParamAppendix}
\scriptsize
\begin{tabular}{|l|l|l|l|l|l|l|l|l|l|l|l|}
\hline
 Model & Grid & hash & LPE & Grid-PEPS & PE & m-PEPS & m-grid & m-hash & m-hashPEPS\\ \hhline{|=|=|=|=|=|=|=|=|=|=|=|}
 Pitted Stonefish& 0.390 & 0.242 & 0.363 & \textbf{0.453} & 0.206 & 0.446 & 0.399 & 0.442 & 0.444\\ \hline 
\textbf{Pitted Stonefish}& 0.456 & 0.350 & 0.450 & \textbf{0.466} & 0.206 & 0.462 & 0.454 & \textbf{0.466} & 0\textbf{.466}\\ \hline 
\end{tabular}
\end{table*}

\section{Conclusion}
Implicit neural representations and positional encoding are some of the most 
efficient methods for learning coordinate to signal functions.
In this paper, we proposed an analysis of the dynamic of 
projecting coordinates into the positional encoding space.
We showed that the motion is of interest as it leads to a unique curve and can help the learning process.
We proposed a generic framework called positional encoding projected sampling (PEPS) 
that tries to use the uniqueness of the positional encoding curve to design a basis for doing learned positional encoding.
This method can be seen as a wrapper around encoders and 
we proposed two implementations of this framework.
The first one; named grid-PEPS,  is a wrapper around a simple grid and strongly improves its learning capabilities in our experimental section.
The second one, named PinkPEPS, is an aggregation strategy based on the power spectral density of natural processing making the process faster and the results more accurate. 
We applied this method to three competitive applications
- implicit image representation, texture compression, and signed distance functions - 
and showed that it was in all the benchmarking better than its counterpart
and reach new state of the art results on quality of rendering.

\bibliographystyle{plain} 
\bibliography{references}

\appendix

\section{Global notes}
\paragraph{Proof of proposition \ref{prop:LissajousUnique} }
Suppose there are two non-null points $(x,y)$ and $(x',y')$ having different ratio $a/b$ and $a'/b'$ and equal Lissajous curve.
As these ratio are different, either $a \neq a'$ or $b \neq b'$.
Consider the first case.
We must have $x \neq x'$.
Yet, if they have the same Lissajous curve, we have
$\sin(x\phi)=\sin(x'\phi)$ for any real $\phi$.
Using $\phi$ as the variable of our sines,
the wave frequency of the left function is $\frac{2\pi}{x}$, 
for the second it is $\frac{2\pi}{x'}$.
For two sine function to be equal they need to have at least the same frequency.
This implies that $\frac{2\pi}{x}=\frac{2\pi}{x'}$,
but $x \neq x'$. This is a contradiction.
The same can be done for $b \neq b'$.
Two points with different ratio cannot have the same Lissajous curve.

We will now focus on points having the same ratio a/b.
Suppose that it exists two points having the same ratio and Lissajous curve.
Those two points can be written by:
$x=nt$,$y=mt$,$x'=nt\alpha$, $y'=mt\alpha$,
with $n,t,\alpha$ a strictly positive number and $\alpha>1$.
Since their Lissajous curve is equal, we have:
$\sin(nt\phi)=\sin(nt\alpha\phi)$ for any strictly positive real $\phi$.
Using $\phi$ as the variable of our sines,
the wave frequency of the left function is $\frac{2\pi}{nt}$, 
for the second it is $\frac{2\pi}{nt\alpha}$.
We should have $\frac{2\pi}{nt}=\frac{2\pi}{nt\alpha}$,
but $nt < nt\alpha$ because $\alpha>1$.
This is a contradiction and the two points must be equal.

\paragraph{No sharing of the grid}
We implemented several version of the grid-PEPS methods,
versions where for each coefficient, a grid was defined and trained, 
and we implemented several strategy of parameters split between grids etc.
This was in most of our experiments dominated by the single shared grid version of PEPS.
One of the reasons for this is that the high frequency grid might miss a lot of the entries of the grid.
This implies that dedicated grid might receive less gradient flow than shared one.

\paragraph{Full LPE}
Learned positional encoding is usually difficult because the largest element might not be present during training.
We performed an ablation study that only uses the APE coordinate and not the original one (we removed point $x$ from Figure~1 from the main paper).
In our results the impact was often marginal, yet always led to slightly less good results.
This implies that PEPS could be an alternative for unseen position in the general case.

\paragraph{Aggregation function}
We tested several aggregation functions instead of the concatenation of PEPS.
We tried to sum all the latent vectors of the PEPS (additive learning).
The results were drastically worse (several order of magnitudes).
We tried to sum them by frequency ($l_{S_i}+l_{C_i}$), results were clearly better than the full sum, 
yet still dominated by the full concatenation one.

\section{Lissajous Curves}
As stated in the main document, even if the rotations of each axis is independent,
linear and simple, 
the multi-dimensional resulting spatial information is more complex.

\begin{figure}[h]
    \centering
    \includegraphics[width=0.35\linewidth]{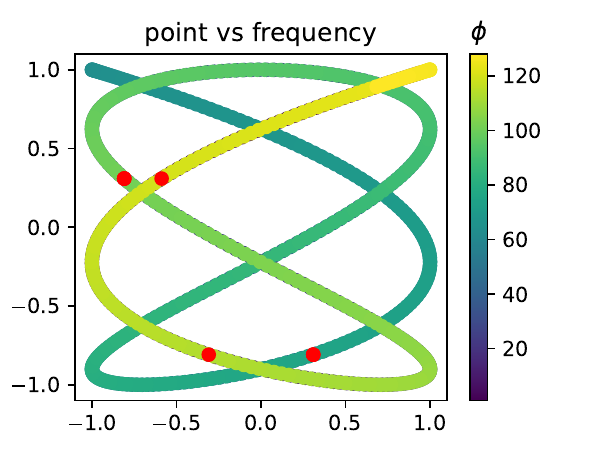}
    \includegraphics[width=0.35\linewidth]{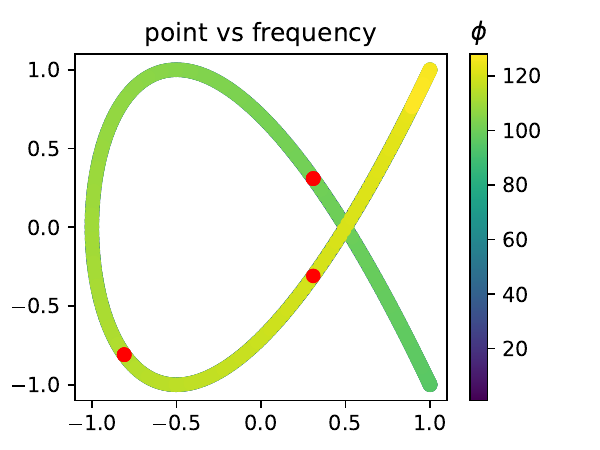}
    \caption{Normalized absolute positional encoding for an interval of frequencies for 
    point (0.35,0.2) on the left and
    point (0.2,0.3) on the right. 
    On both curves, red dots are the points at frequencies $2^i\pi$ for i in $\{1,2,3,4\}$.}
    \label{fig:Lissajous}
\end{figure}

Consider first the sinusoidal 2D case.
Given a point $(x,y)$, and a frequency $\phi_i$,
the generated point is $(\sin{\phi_i x},\sin{\phi_i y})$.
This is a particular case of Lissajous curve.
Lissajous curves \cite{lissajous1857memoire},
from the physicists Jules Antoine Lissajous,
are considered a particular case of complex harmonic motions and often used
in aerial search and polynomials interpolation \cite{steckenrider2024lissajous,erb2016bivariate2}.
The Lissajous curve definition is given by the parametric equations:
\begin{align}\label{eq:Lissajous}
    x = A \sin at+\delta, y = B \sin bt
\end{align}
It is known for describing two perpendicular oscillations in $x$ and $y$ directions, 
and at different angular frequencies (i.e., $a$ and $b$).
Obviously, the positional encoding of a given point $(x,y)$ will follow a Lissajous
curve when the frequency will vary.
Figure~\ref{fig:Lissajous} is an example of two points and their associated Lissajous curve.
The two important parameters of Lissajous curves are the frequency ratio $a/b$ and the shift noted $\delta$.
The $a/b$ ratio determines the number of "lobes" of the figure. 
As you can see, in figure~\ref{fig:Lissajous}, the left figures makes 7 horizontal pass and 4 vertical ones.
The right image makes 2 horizontal and 3 verticals.
As this is a "counting", the ratio must be rational for the figure to be close.
Finally, the $\delta$ parameter manage the shifting of the curve.
As shown in the figure and the next paragraph, 
this $\delta$ is always equal to zero when considering the even positional encoding and also fixed for the odd one, so it can be discarded from our analysis.

\begin{figure}
  \centering
  \includegraphics[width=0.45\linewidth]{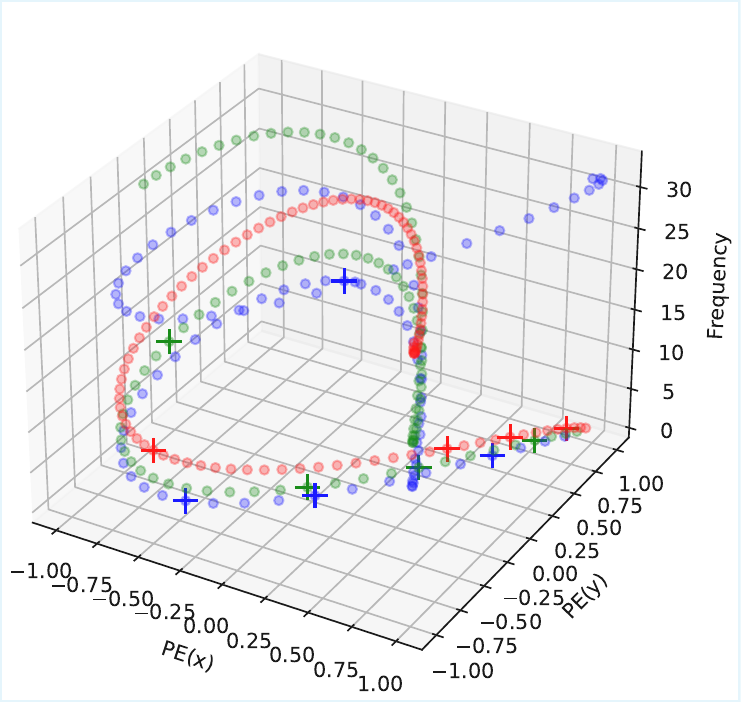}
  \caption{Normalized absolute positional encoding for an interval of frequencies for 
    points (0.2,0.3) in red, (0.3,0.45)  in green and (0.4,0.6) in blue.
    All those points have the same ratio and no difference in delta, yet they
    result in different curves.
    The cross (+) points are sampled at equivalent frequencies.
    We can see that they result is clearly different positions.}
    \label{fig:LissajousUnique}
\end{figure}

\paragraph{Discrete coordinate case}
Consider for example the case where the PE is applied to 2D textures.
Each pixel location is given using an equally spaced coordinate system.
Coordinates of this system are of the form $x = nt, y = mt$.
where $t$ is the increment size (for example the pixel size).
Given that each point of the original texture can be mapped to this coordinate,
any sampled point will have a rational ratio $\frac{n}{m}$
and no $\delta$.
Thanks to these, the Lissajous curve of each of the point of this 
coordinate system will be close and the curve is well-defined.
Note that Lissajous curves have the same properties 
for more than 2 dimensions \cite{dencker2017multivariate}.
The closeness property still hold in higher dimensional spaces if 
the ratio of each pair of coordinates is rational.

\section{Image Representation}
In the paper, the use of a L1 loss, of a fixed learning rate and a simple leaky ReLU activation function was to ensure
that the variability in the results was not a product of several potentially unstable components.
In addition to the results in the paper, we tested our encoders using
MLP having a GeLU activation function, two distinct learning rates (0.001 for the MLP, and 0.1 for the grids)
and a cosine adaptative learning rate.
The impact was that all methods have been improved, but the global order remain the same.
Table~\ref{tab:ntckodakL2} shows the difference in results for the grid and ntc PEPS method for the kodak dataset
Yet those adaptive learning rates and various activation functions often led to instability.
Another example is the use of the WIRE activation function from \cite{saragadam2023wire}.
Those led to large instability and forced decrease in the learning rate, 
and most importantly strictly longer training, for marginal gains.
This prevent fair comparison.
That is the reason why the paper is focusing on "vanilla" MLPs to ensure the stability of the result and the measurement of the impact of the encoder.

\begin{table}[h!]
\centering
\caption{Impact of the loss, architecture, and learning rate for the Kodak dataset.}
\label{tab:ntckodakL2}
\begin{tabular}{|l|c|c|c|}
\hline
Metric  & Grid & G-P-PEPS & NTC$_{N}$\\  \hhline{|=|=|=|=|}
PSNR L1 ($\uparrow$)& 40.871 & 44.237 & 41.229 \\ \hline
PSNR L2 ($\uparrow$)& 45.30 & 47.83 & 44.87 \\ \hhline{|=|=|=|=|}
SSIM L1 ($\uparrow$)& 0.973 & 0.975 & 0.968 \\ \hline
SSIM L2 ($\uparrow$)& 0.992 & 0.993 & 0.992 \\ \hline
\end{tabular}
\end{table}
\subsection{Metrics}

\paragraph{Compression quality: PSNR.}
Peak signal-to-noise ratio (PSNR) is a well-known metric used for the measurement of quality of lossy compression technics.
It is defined by $PSNR(I,\hat{I}) = 20\log_{10}(Max(I))-10\log_{10}(MSE(I,\hat{I}))$,
where $MSE$ is the mean squared error. 

\paragraph{Perceptual quality: SSIM and FLIP.}
Structural Similarity Index Measure (SSIM) is a perception based measure trying to characterize of the degradation of the image. 
For a complete definition, please refer to \cite{wang2004image}.
In our implementation, we used the \textit{StructuralSimilarityIndexMeasure} from the \textit{torchmetrics.image} package. 
For the flip error \cite{andersson2020flip}, we used the \textit{flip\_evaluator} python packaged.
The FLIP error reported in this paper is the average of the FLIP error of an image.

\paragraph{Spectral quality: LSD and LSPD.}
In this paper, we introduce the use of the two spectral metrics
log-spectral distance (LSD) and log-power spectral distance (LPSD).
Intuitively, the first one is an analysis of the spectral quality directly through its 2D Fourier transform.
This measure can be seen as a global structure measure done through the average amplitude distance between phases.
The second one is a direct measurement of the distance in power spectral density. 
While it is known that natural images follows a $1/f^\alpha$ PSD law, 
the LPSD quantify the distance between this law and the learned image.
More precisely. the 2D Fourier transform of the image is 2 dimensionsial.
Thos two dimensions are aggregated by taking the average statistics over all coordinate vectors whose norm 2 are equals.
More precisely  $||f||_2=||f'||_2$ for $f,f' \in \mathbb{R}^2$ with the norm 2
defined by $||f||_2\sqrt{f_1^2+f_2^2}$.
The resulting graph is for example the one showed in Figure~\ref{img:energy}.
Then the LPSD is simply the L1 distance between the original image and the learned one.
More information about the PSD or those metrics can be found in \cite{van1996modelling,field1987relations,rabiner1993fundamentals,gray2003distance}

\section{NTC}
\paragraph{Dataset}
We used 18 instances from https://polyhaven.com/ and https://ambientcg.com/. 
We extracted the available one from \cite{fujieda2024neural,ubi24} and some more.
The complete list of instances is: 
bench vice 01, cardboard box 01, cannon 01, clay roof tiles 02,
fabric pattern 07, garden gnome, garden sprinkler 01, wood planks,
treasure chest, Paving Stones 070, Rails 001, red dirt mud 01,
Aerial Rocks 02, Bricks 090, forest sand 01, Metal Plates 013, roof 09, Wood 063

\paragraph{Methods}
The methods used in the NTC section of the paper are:
\begin{itemize}
    \item BI grid. A basic grid of resolution $r$ having in each discrete coordinate from $[0,r-1]^2$ a latent vector of dimension $d$. When an input coordinate is given this method extract the four closest latent vectors in the grid and applies a bilinear interpolation between them. The result is of dimension $d$.
    \item LPE. Local Positional Encoding \cite{Fujieda23local}, is a post process of the BI grid. It multiplies the result of the BI grid with the \textit{local} positional encoding of the input. The local positional encoding being the positional encoding with respect to the cell of the grid used by the BI grid
    \item NTC$_{N}$. The neural texture compression \cite{nvidNTC23} is composed of 2 grids. The first one is a concatenation grid, which implies that the four closest latent vectors are returned by the grid. The second grid is a BI grid. The last element is the Tiled positional encoding, which correspong to the last 3 frequencies of the PE.
\end{itemize}
Two distinct learning rates (0.001 for the MLP, and 0.1 for the grids)
and a cosine adaptative learning rate. 
the activation function of the network are Gelu for fair comparison with \cite{nvidNTC23},
batch size is 60k, and 3k epochs of 40 batchs are done.
For each batch, a random set of coordinate is sampled from pixel locations.

\paragraph{Results}
We put in the supplementary random batches of samples from the 4K texture sets.
The complete set of 4K results is about 10 GB, and cannot be part of the supplementary.

\section{SDF}
The scan of Pitted Stonefish is available at \cite{ffishAsiaErosa2023}. 
Other instances are from  \cite{Fujieda23local}.
They are converted to SDFs using our application written in C++ and HIP.
All rendering are available in the multimedia supplementary.
We provide in this document two results for the L1 loss.
The learning rate is 0.001, 
the activation function of the network are SILU,
batch size is 60k, and 3k epochs of 40 batchs are done.
For each batch, a random set of coordinate is sampled from $[0,1]^3$.
The bold line for stonefish is using 8 times more parameters for the encoder.

\paragraph{MAPE} Mean Absolute Percentage Error (MAPE) 
is a loss that measures the average magnitude of error in a set. 
It is defined by:
$$
MAPE(p, gt) = 100\frac{1}{n}\sum^n_i\frac{|gt-p|}{|gt|}
$$

Results for the four SDF instances, using the same setting of the paper and the L1 loss instead of MAPE loss are given in table \ref{tab:SDFAppendix}.
The only difference is the learning rate which is set to 0.01.
As we can see, just like for images, the PEPS methods remain the best, and especially grid-PEPS that is clearly above.

\begin{table*}[ht]
\centering
\caption{(SDF) IoU ($\uparrow$) results for all the methods using the L1 loss.}
\label{tab:SDFAppendix}
\scriptsize
\begin{tabular}{|l|l|l|l|l|l|l|l|l|l|l|l|}
\hline
 Model & Grid & hash & LPE & Grid-PEPS & PE & m-PEPS & m-grid & m-hash & m-hashPEPS\\ \hhline{|=|=|=|=|=|=|=|=|=|=|=|}
 
 Lucy & 0.906 & 0.779 & 0.904 & \textbf{0.909} & 0.874 & \textbf{0.909} & 0.906 & \textbf{0.909} & \textbf{0.909}\\ \hline 
 Pitted Stonefish& 0.390 & 0.242 & 0.363 & \textbf{0.453} & 0.206 & 0.446 & 0.399 & 0.442 & 0.444\\ \hline 
 Thai Statue & 0.927 & 0.747 & 0.925 & \textbf{0.930} & 0.900 & \textbf{0.930} & 0.928 & \textbf{0.930} & \textbf{0.930}\\ \hline 
 Armadillo & 0.956 & 0.761 & 0.956 & \textbf{0.957} & 0.943 & \textbf{0.957} & \textbf{0.957} & \textbf{0.957} & \textbf{0.957}\\ \hhline{|=|=|=|=|=|=|=|=|=|=|=|}
 
 global & 0.795 & 0.632 & 0.787 & \textbf{0.812} & 0.731 & \textit{0.810} & 0.798 & \textit{0.810} & \textit{0.810}\\ \hline
\end{tabular}
\end{table*}

\end{document}